\documentclass{article}

\usepackage{microtype}
\usepackage{graphicx}
\usepackage{subcaption}
\usepackage{booktabs} %
\usepackage[hyphens]{url}
\graphicspath{{figures/}}
 \usepackage{color}
\usepackage{hyperref}

\usepackage[accepted]{icml2019}

\icmltitlerunning{We Need No Pixels: Video Manipulation Detection Using Stream Descriptors}

\begin{document}

\twocolumn[
\icmltitle{We Need No Pixels: Video Manipulation Detection Using Stream Descriptors}

\begin{icmlauthorlist}
\icmlauthor{David G\"{u}era}{pu}
\icmlauthor{Sriram Baireddy}{pu}
\icmlauthor{Paolo Bestagini}{mi}
\icmlauthor{Stefano Tubaro}{mi}
\icmlauthor{Edward J. Delp}{pu}
\end{icmlauthorlist}

\icmlaffiliation{pu}{Video and Image Processing Laboratory (VIPER), Purdue University, West Lafayette, Indiana, USA}
\icmlaffiliation{mi}{Dipartimento di Informazione, Elettronica e Bioingegneria, Politecnico di Milano, Milano, Italy}

\icmlcorrespondingauthor{David G\"{u}era}{dgueraco@purdue.edu}
\icmlcorrespondingauthor{Edward J. Delp}{ace@ecn.purdue.edu}

\icmlkeywords{Machine Learning, ICML, Video Manipulations, Deepfakes, Forensics,Detection, MediFor}

\vskip 0.3in
]

\printAffiliationsAndNotice%

\begin{abstract}
Manipulating video content is easier than ever.
Due to the misuse potential of manipulated content, multiple detection techniques that analyze the pixel data from the videos have been proposed.
However, clever manipulators should also carefully forge the metadata and auxiliary header information, which is harder to do for videos than images.
In this paper, we propose to identify forged videos by analyzing their multimedia stream descriptors with simple binary classifiers, completely avoiding the pixel space.
Using well-known datasets, our results show that this scalable approach can achieve a high manipulation detection score if the manipulators have not done a careful data sanitization of the multimedia stream descriptors.

\end{abstract}

\section{Introduction}
\label{intro}
Video manipulation is now within reach of any individual.
Recent improvements in the machine learning field have enabled the creation of powerful video manipulation tools.
Face2Face~\cite{Thies2016}, Recycle-GAN~\cite{Bansal2018}, Deepfakes~\cite{Korshunov2018}, and other face swapping techniques~\cite{Korshunova2017} embody the latest generation of these open source video forging methods.
It is assumed as a certainty both by the research community~\cite{Brundage2018} and governments across the globe~\cite{Vincent2018, Chesney2018} that more complex tools will appear in the near future.
Classical and current video editing methods have already demonstrated dangerous potential, having been used to generate political propaganda~\cite{Bird2015}, revenge-porn~\cite{Curtis2018}, and child-exploitation material~\cite{Cole2018}.

\begin{figure}[t]
    \begin{center}
    \centerline{\includegraphics[width=\columnwidth]{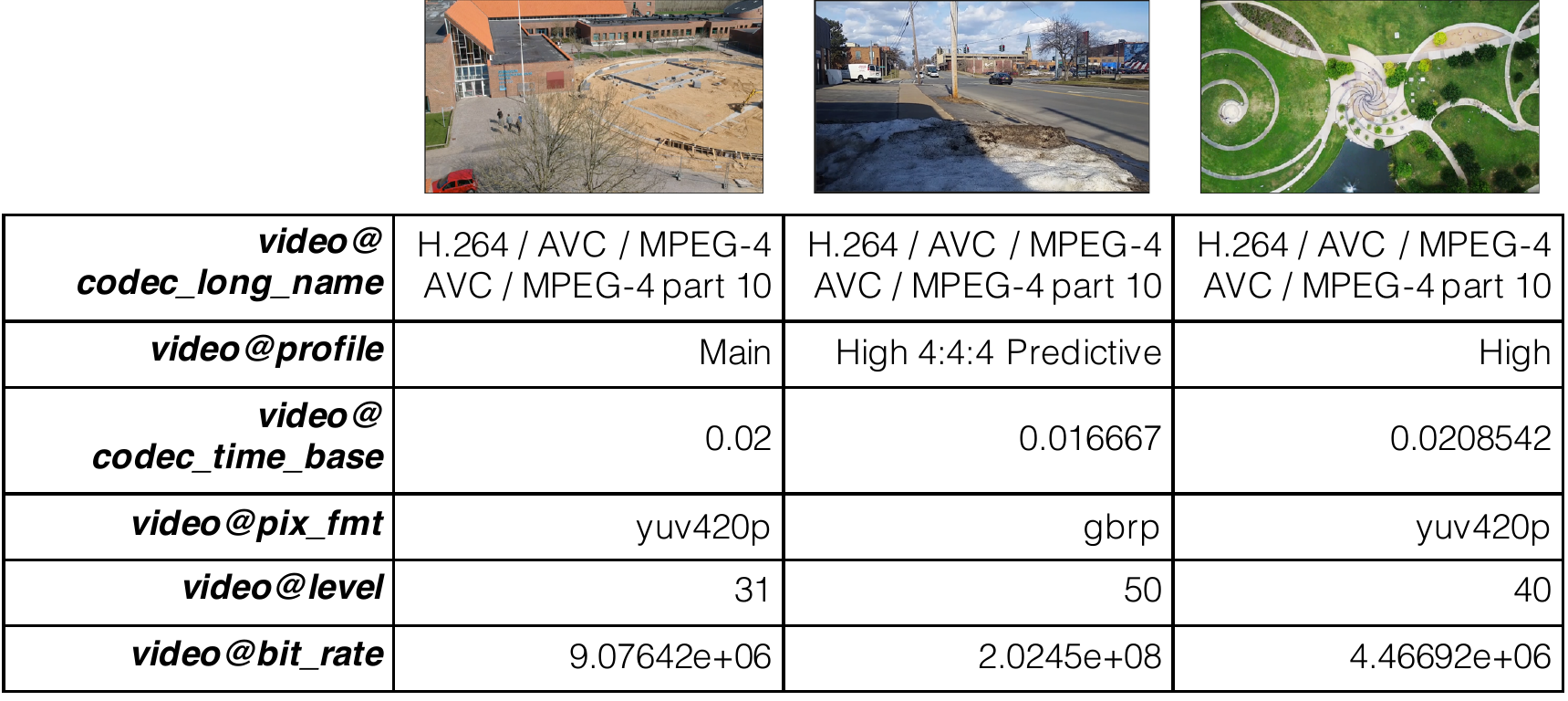}}
    \caption{Examples of some of the information extracted from the video stream descriptors. These descriptors are necessary to decode and playback a video.}
    \label{fig:icml-historical}
    \end{center}
    \vskip -0.3in
\end{figure}

Due to the ever increasing sophistication of these techniques, uncovering manipulations in videos remains an open problem.
Existing video manipulation detection solutions focus entirely on the observance of anomalies in the pixel domain of the video.
Unfortunately, it can be easily seen from a game theoretic perspective that, if both manipulators and detectors are equally powerful, a Nash equilibrium will be reached~\cite{Stamm2012}.
Under that scenario, both real and manipulated videos will be indistinguishable from each other, and the best detector will only be capable of random guessing. %
Hence, methods that look beyond the pixel domain are critically needed.
So far, little attention has been paid to the necessary metadata and auxiliary header information that is embedded in every video.
As we shall present, this information can be exploited to uncover unskilled video content manipulators. 

In this paper, we introduce a new approach to address the video manipulation detection problem.
To avoid the zero-sum, leader-follower game that characterizes current detection solutions, our approach completely avoids the pixel domain.  
Instead, we use the multimedia stream descriptors~\cite{jack2007_ch13} that ensure the playback of any video (as shown in Figure~\ref{fig:icml-historical}).
First, we construct a feature vector with all the descriptor information for a given video.
Using a database of known manipulated videos, we train an ensemble of  a support vector machine and a random forest that acts as our detector.
Finally, during testing, we generate the feature vector from the stream descriptors of the video under analysis, feed it to the ensemble, and report a manipulation probability. 

The contributions of this paper are summarized as follows.
First, we introduce a new technique that does not require access to the pixel content of the video, making it fast and scalable, even on consumer grade computing equipment. 
Instead, we rely on the multimedia descriptors present on any video, which are considerably harder to manipulate due to their role in the decoding phase.
Second, we thoroughly test our approach using the NIST MFC datasets~\cite{Guan2019} and show that even with a limited amount of labeled videos, simple machine learning ensembles can be highly effective detectors.
Finally, all of our code and trained classifiers will be made available\footnote{\url{https://github.com/dguera/fake-video-detection-without-pixels}} so the research community can reproduce our work with their own datasets.

\section{Related Work}
\label{works}
The multimedia forensics research community has a long history of trying to address the problem of detecting manipulations in video sequences. 
\cite{Milani2012} provide an extensive and thorough overview of the main research directions and solutions that have been explored in the last decade.
More recent work has focused on specific video manipulations, such as local tampering detection in video sequences~\cite{Stamm2012b, Bestagini2013}, video re-encoding detection~\cite{Bian2014, Bestagini2016}, splicing detection in videos~\cite{Hsu2008, Mullan2017, Mandelli2018}, and near-duplicate video detection~\cite{Bayram2008, Lameri2017}.
\cite{Amiano2015,Amiano2019} also present solutions that use 3D PatchMatch~\cite{Barnes2009} for video forgery detection and localization, whereas~\cite{Avino2017} suggest using data-driven machine learning based approaches.
Solutions tailored to detecting the latest video manipulation techniques have also been recently presented.
These include the works of \cite{Li2018,Guera2018b} on detecting Deepfakes and \cite{roessler2018,Matern2018} on Face2Face~\cite{Thies2016} manipulation detection.

As covered by \cite{Milani2012}, image-based forensics techniques that leverage camera noise residuals~\cite{Khanna2008}, image compression artifacts~\cite{Bianchi2012}, or geometric and physics inconsistencies in the scene~\cite{Bulan2009} can also be used in videos when applied frame by frame.
In~\cite{Fan2011} and~\cite{Huh2018}, Exif image metadata is used to detect either image brightness and contrast adjustments, and splicing manipulations in images, respectively.
Finally,~\cite{Iuliani2019} use video file container metadata for video integrity verification and source device identification.
To the best of our knowledge, video manipulation detection techniques that exploit the multimedia stream descriptors have not been previously proposed.
\section{Proposed Method}
\label{method}
\begin{figure*}[t]
    \centering
    \begin{subfigure}[b]{1\textwidth}
       \centering
       \includegraphics[width=1\linewidth]{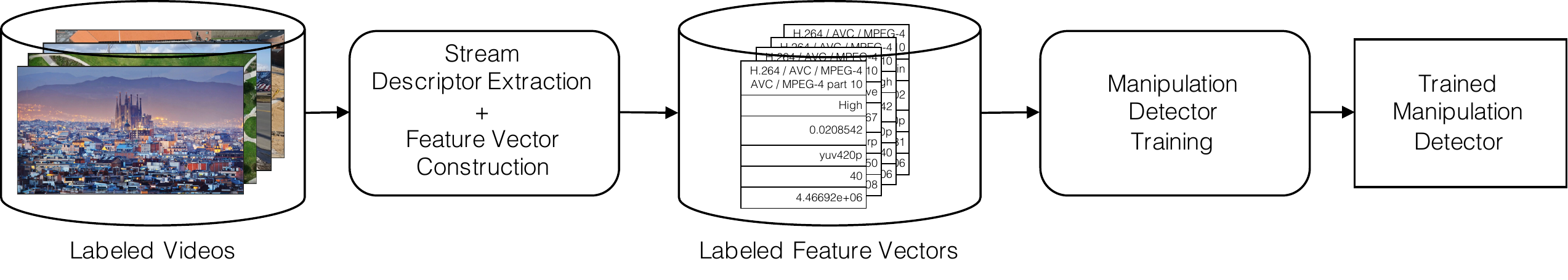}
       \caption{}
       \label{fig:method_train} 
    \end{subfigure}
    \vskip 0.1in

    \begin{subfigure}[b]{1\textwidth}
       \centering
       \includegraphics[width=0.8\linewidth]{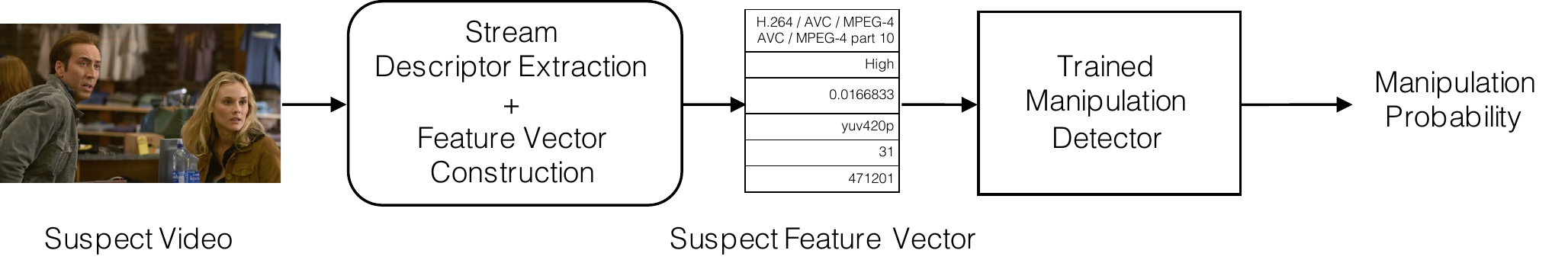}
       \caption{}
       \label{fig:method_test}
    \end{subfigure}
    \vskip -0.1in

    \caption[Two stage approach]{(a) Block diagram of the training stage of our proposed method. We process a labeled database of manipulated and pristine videos to generate a feature vector for each video from its multimedia stream descriptors. These feature vectors are then used to train and select the best detector (b) Block diagram of the testing stage of our proposed method. Given a suspect video, a feature vector is generated and processed by the previously selected detector. Finally, a manipulation probability for the suspect video is reported.}
    \label{fig:method}
    \vskip -0.1in
\end{figure*}

Current video manipulation detection approaches rely on uncovering manipulations by studying pixel domain anomalies.
Instead, we propose to use the multimedia stream descriptors of videos as our main source of information to spot manipulated content.
To do so, our method works in two stages, as presented in Figure~\ref{fig:method}.
First, during the training phase, we extract the multimedia stream descriptors from a labeled database of manipulated and pristine videos. 
In practice, such a database can be easily constructed using a limited amount of manually labeled data coupled with a semi-supervised learning approach, as done by~\cite{Zannettou2018}.
Then, we encode these descriptors as a feature vector for each given video.
We apply median normalization to all numerical features.
As for categorical features, each is encoded as its own unique numerical value.
Once we have processed all the videos in the database, we use all the feature vectors to train different binary classifiers as our detectors.
More specifically, we use a random forest, a support vector machine (SVM) and an ensemble of both detectors.
The best hyperparameters for each detector are selected by performing a random search cross-validation over a $10$-split stratified shuffling of the data and $1{,}000$ trials per split. 
Figure~\ref{fig:method_train} summarizes this first stage.
In our implementation, we use \textit{ffprobe}~\cite{Bellard2019} for the multimedia stream descriptor extraction.
For the encoding of the descriptors as feature vectors, we use \textit{pandas}~\cite{Mckinney2010} and \textit{scikit-learn}~\cite{Pedregosa2011}. 
As for the training and testing of the SVM, the random forest, and the ensemble, we use the implementations available in the~\textit{scikit-learn} library.

Figure~\ref{fig:method_test} shows how our method would work in practice.
Given a suspect video, we extract its stream descriptors and generate its corresponding feature vector, which is normalized based on the values learnt during the training phase.
Since some of the descriptor fields are optional, we perform additional post-processing to ensure that the feature vector can be processed by our trained detector.
Concretely, if any field is missing in the video stream descriptors, we perform data imputation by mapping missing fields to a fixed numerical value.
If previously unseen descriptor fields are present in the suspect video stream, they are ignored and not included in the corresponding suspect feature vector.
Finally, the trained detector analyzes the suspect feature vector and computes a manipulation probability.

It is important to note that although our approach may be vulnerable to video re-encoding attacks, this is traded off for scalability, a limited need of labeled data, and a high video manipulation detection score, as we present in Section~\ref{experiments}.
Also, the fact that our solution is orthogonal to pixel-based methods and requires limited amounts of data, which means that ideally, we could use both approaches simultaneously.
Our approach could be used to quickly identify manipulated videos, minimizing the need to rely on human annotation.
Later, these newly labeled videos could be used to improve the performance of pixel-based video manipulation detectors.
Finally, following the recommendations of~\cite{Brundage2018}, we want to reflect on a potential misuse of the proposed approach.
We believe that our approach could be misused by someone with access to large amounts of labeled video data.
Using that information, a malevolent adversary could identify specific individuals, such as journalists or confidential informants, who may submit anonymous videos using the same devices they use to upload videos to social media websites.
To avoid this, different physical devices or proper video data sanitization should be used.

\section{Experimental Results}
\label{experiments}
\subsection{Datasets}
\label{dataset}
In order to evaluate the performance of our proposed approach, we use the Media Forensics Challenge (MFC) datasets~\cite{Guan2019}. 
Collected by the National Institute of Standards and Technology (NIST), this data comprises over $11{,}000$ high provenance videos and $4{,}000$ manipulated videos.
In our experiments, we use the videos from the following datasets for training, hyper-parameter selection, and validation: the Nimble Challenge $2017$ development dataset, the MFC18 development version 1 and version 2 datasets, and the MFC18 GAN dataset.
This represents a total of $677$ videos, of which $167$ are manipulated. 
For testing our model, we use the MFC18 evaluation dataset and the MFC19 validation dataset, which have a total of $1{,}097$ videos. Of those videos, $336$  have been manipulated.

\subsection{Experimental Setup}
\label{setup}
To show the merits of our method in terms of scalability and limited compute requirements, we design the following experiment.
First, we select machine learning binary classifiers that are well known for their modeling capabilities, even with limited access to training samples.
As previously mentioned, we use a random forest, a support vector machine, and a soft voting classification ensemble with both. 
This final ensemble is weighted 4 to 1 in favor of the decision of the random forest.
Then, to show the performance of each detector under different data availability scenarios, we train them using $10$\%, $25$\%, $50$\% and $75$\% of the available training data. 
We use a stratified shuffle splitting policy to select these training subsets, meaning that the global ratio of manipulated to non-manipulated videos of the entire training set is preserved in the subsets.
In all scenarios, a sequestered 25\% subset of the training data is used for hyper-parameter selection and validation.
Finally, the best validated model is selected for testing.
Due to the imbalance of manipulated to non-manipulated videos, we use the Precision-Recall (PR) curve as our evaluation metric, as recommended by~\cite{Saito2015}.
We also report the F1 score, the area under the curve (AUC) score, and the average precision (AP) score for each classifier.

\subsection{Results and Discussion}
\label{results}
\begin{figure}[t]
    \centering
    \includegraphics[width=1\linewidth]{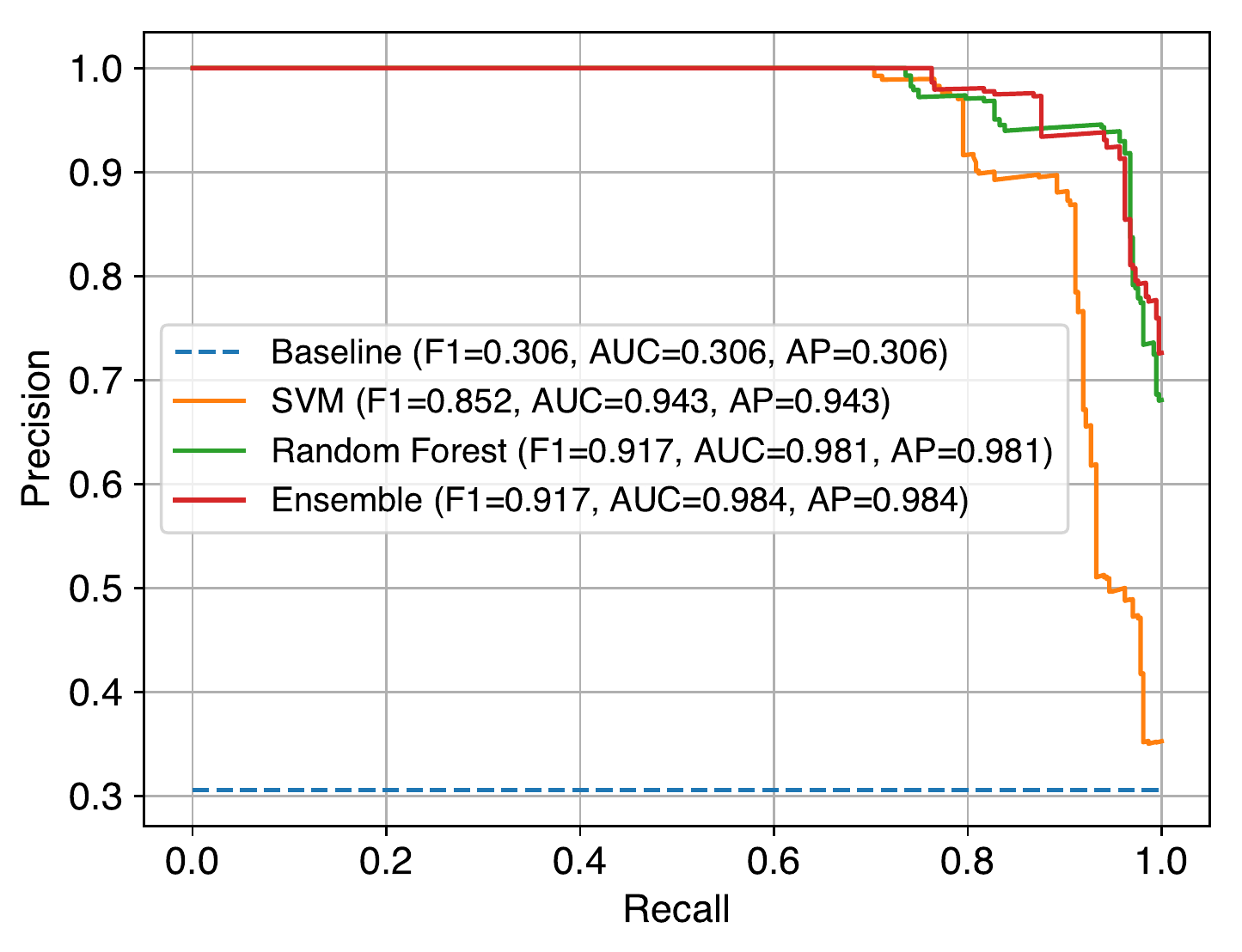}
    \vskip -0.2in
    \caption{PR curves, F1 score, AUC score, and AP score on the test set for all the trained models using $10$\% of the available training data (68 videos).}
    \label{fig:exp_10}
    \vskip -0.1in
\end{figure}

\begin{figure}[t]
    \centering
    \includegraphics[width=1\linewidth]{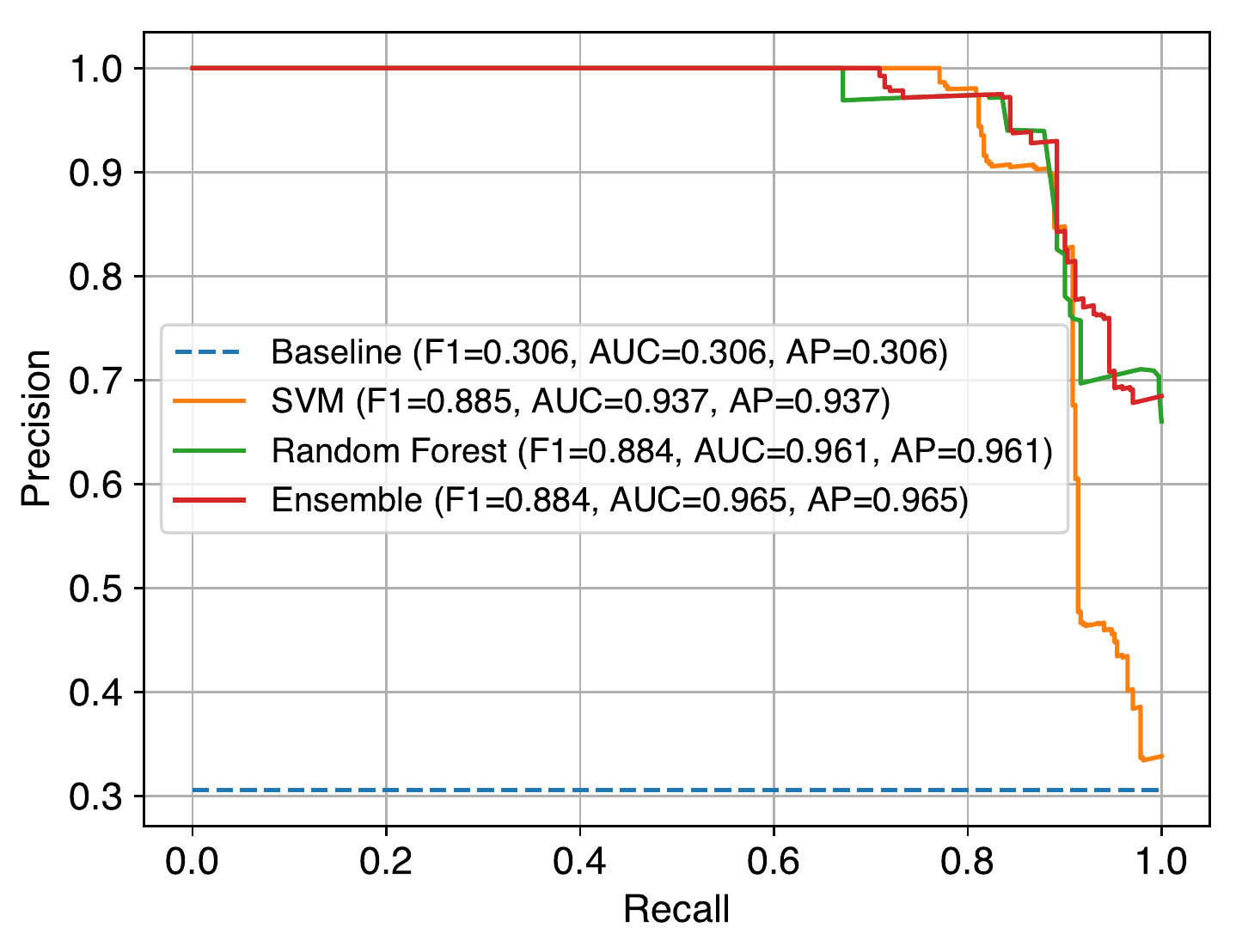}
    \vskip -0.2in
    \caption{PR curves, F1 score, AUC score, and AP score on the test set for all the trained models using $25$\% of the available training data (169 videos).}
    \label{fig:exp_25}
    \vskip -0.1in
\end{figure}

\begin{figure}[t]
    \centering
    \includegraphics[width=1\linewidth]{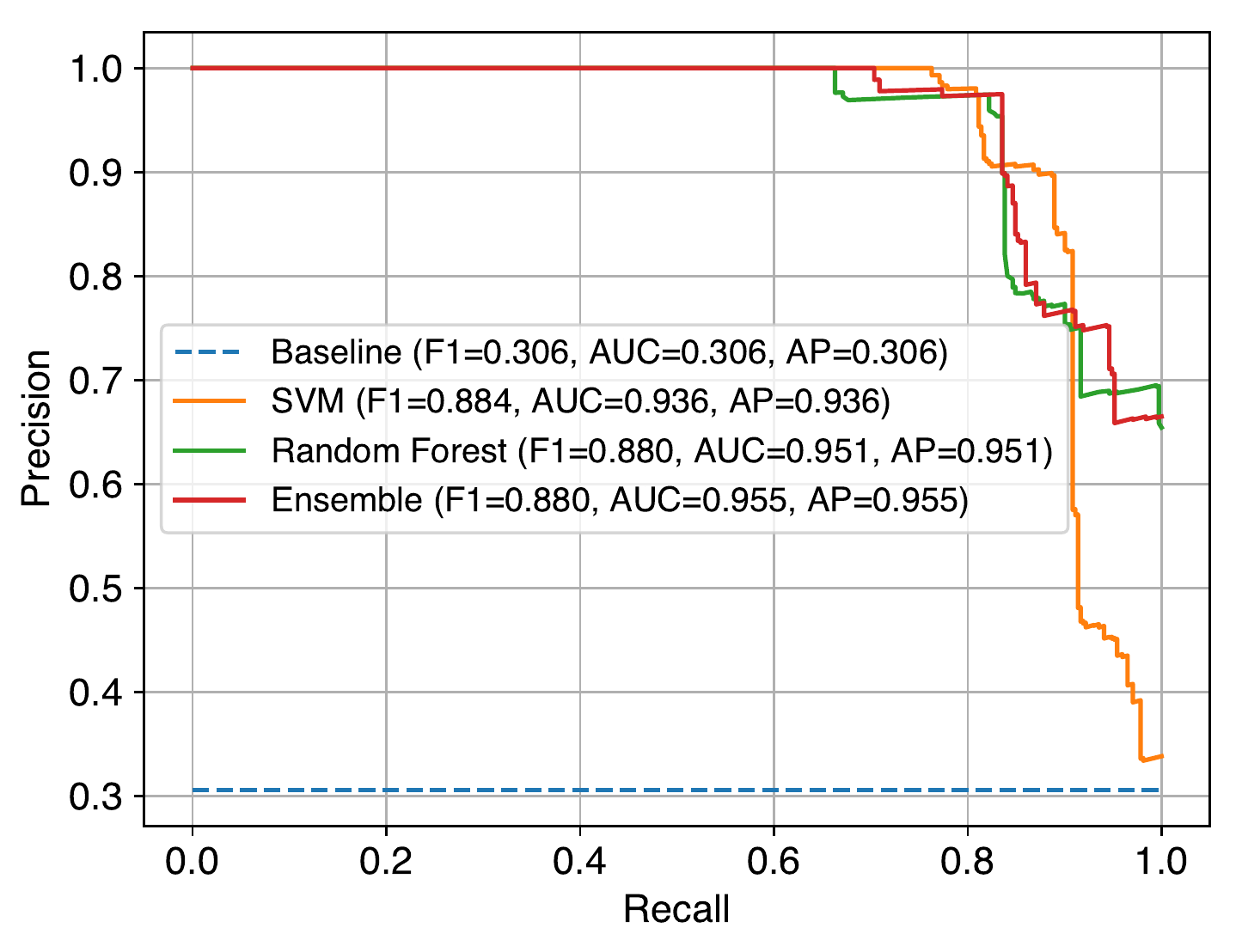}
    \vskip -0.2in
    \caption{PR curves, F1 score, AUC score, and AP score on the test set for all the trained models using $50$\% of the available training data (339 videos).}
    \label{fig:exp_50}
    \vskip -0.1in
\end{figure}

\begin{figure}[t]
    \centering
    \includegraphics[width=1\linewidth]{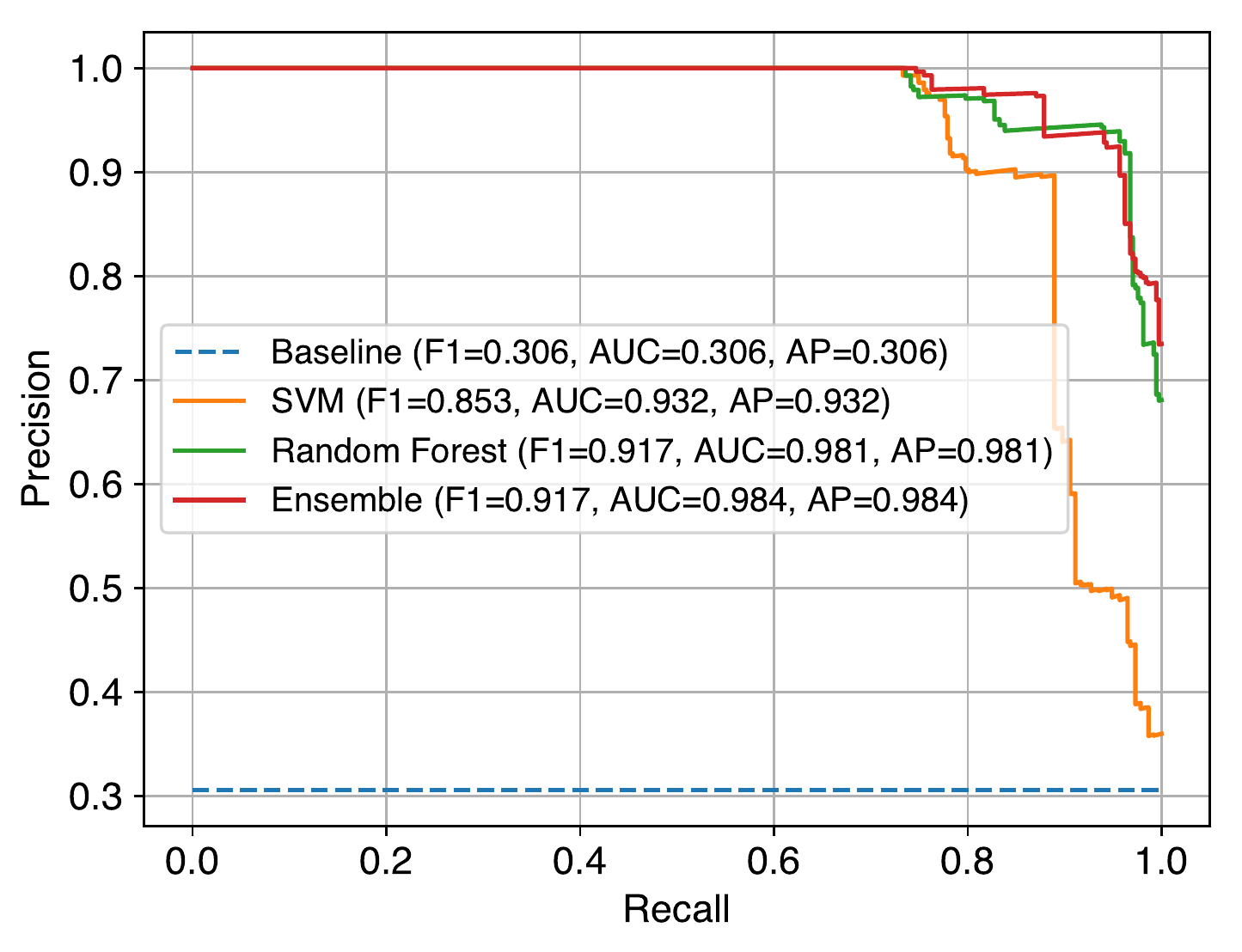}
    \vskip -0.2in
    \caption{PR curves, F1 score, AUC score, and AP score on the test set for all the trained models using $75$\% of the available training data (508 videos).}
    \label{fig:exp_75}
    \vskip -0.1in
\end{figure}

As we can see in Figure~\ref{fig:exp_10}, Figure~\ref{fig:exp_25}, Figure~\ref{fig:exp_50}, and Figure~\ref{fig:exp_75}, under all scenarios the voting ensemble of the random forest and the support vector machine generally achieves the best overall results, followed by the random forest and the SVM.
More specifically, our best ensemble model achieves a F1 score of $0.917$, an AUC score of $0.984$ and an AP score of $0.984$.
To contextualize these results, we have included the performance of a binary classifier baseline which predicts a video manipulation with probability $p=0.306$. 
This corresponds to the true fraction of manipulated videos in the test set. 
Note that it is higher than the fraction of manipulated videos in the training subsets, which is $0.247$.
This baseline model would achieve an F1, AUC, and AP score of $0.306$.
We can see that our best model is three times better than the baseline in all reported metrics.
Notice that, as seen in Figure~\ref{fig:exp_10}, the ensemble trained with $68$ videos has achieved equal or better results than the ensembles trained with more videos.
This shows that, even with a very limited number of stream descriptors, a properly tuned machine learning model can be trained easily to spot video manipulations.


\section{Conclusion}
\label{conclusion}
Up until now, most video manipulation detection techniques have focused on analyzing the pixel data to spot forged content.
In this paper, we have shown how simple machine learning classifiers can be highly effective at detecting video manipulations when the appropriate data is used.
More specifically, we use an ensemble of a random forest and an SVM trained on multimedia stream descriptors from both forged and pristine videos.
With this approach, we have achieved an extremely high video manipulation detection score while requiring very limited amounts of data.
Based on our findings, our future work will focus on techniques that automatically perform data sanitization.
This will allow us to remove metadata and auxiliary header information that may give away sensitive information such as the source of the video.


\section*{Acknowledgements}
\label{ack}
This material is based on research sponsored by DARPA and Air Force Research Laboratory (AFRL) under agreement number FA8750-16-2-0173. 
The U.S. Government is authorized to reproduce and distribute reprints for Governmental purposes notwithstanding any copyright notation thereon. 
The views and conclusions contained herein are those of the authors and should not be interpreted as necessarily representing the official policies or endorsements, either expressed or implied, of DARPA and Air Force Research Laboratory (AFRL) or the U.S. Government. 

\bibliography{references}

\begin{thebibliography}{42}
\providecommand{\natexlab}[1]{#1}
\providecommand{\url}[1]{\texttt{#1}}
\expandafter\ifx\csname urlstyle\endcsname\relax
  \providecommand{\doi}[1]{doi: #1}\else
  \providecommand{\doi}{doi: \begingroup \urlstyle{rm}\Url}\fi

\bibitem[Bansal et~al.(2018)Bansal, Ma, Ramanan, and Sheikh]{Bansal2018}
Bansal, A., Ma, S., Ramanan, D., and Sheikh, Y.
\newblock Recycle-{GAN}: Unsupervised video retargeting.
\newblock \emph{Proceedings of the European Conference on Computer Vision},
  pp.\  119--135, September 2018.
\newblock URL \url{https://doi.org/10.1007/978-3-030-01228-1_8}.
\newblock {Munich, Germany}.

\bibitem[Barnes et~al.(2009)Barnes, Shechtman, Finkelstein, and
  Goldman]{Barnes2009}
Barnes, C., Shechtman, E., Finkelstein, A., and Goldman, D.~B.
\newblock Patchmatch: A randomized correspondence algorithm for structural
  image editing.
\newblock \emph{ACM Transactions on Graphics}, 28\penalty0 (3):\penalty0
  24:1--24:11, July 2009.
\newblock URL \url{https://doi.org/10.1145/1531326.1531330}.

\bibitem[Bayram et~al.(2008)Bayram, Sencar, and Memon]{Bayram2008}
Bayram, S., Sencar, H.~T., and Memon, N.
\newblock Video copy detection based on source device characteristics: A
  complementary approach to content-based methods.
\newblock \emph{Proceedings of the ACM International Conference on Multimedia
  Information Retrieval}, pp.\  435--442, October 2008.
\newblock URL \url{https://doi.org/10.1145/1460096.1460167}.
\newblock Vancouver, British Columbia, Canada.

\bibitem[Bellard et~al.(2019)]{Bellard2019}
Bellard, F. et~al.
\newblock ffprobe documentation.
\newblock April 2019.
\newblock URL \url{https://www.ffmpeg.org/ffprobe.html}.
\newblock (Accessed on 04/17/2019).

\bibitem[Bestagini et~al.(2013)Bestagini, Milani, Tagliasacchi, and
  Tubaro]{Bestagini2013}
Bestagini, P., Milani, S., Tagliasacchi, M., and Tubaro, S.
\newblock Local tampering detection in video sequences.
\newblock \emph{Proceedings of the IEEE International Workshop on Multimedia
  Signal Processing}, pp.\  488--493, September 2013.
\newblock URL \url{https://doi.org/10.1109/MMSP.2013.6659337}.
\newblock {Pula, Italy}.

\bibitem[{Bestagini} et~al.(2016){Bestagini}, {Milani}, {Tagliasacchi}, and
  {Tubaro}]{Bestagini2016}
{Bestagini}, P., {Milani}, S., {Tagliasacchi}, M., and {Tubaro}, S.
\newblock Codec and gop identification in double compressed videos.
\newblock \emph{IEEE Transactions on Image Processing}, 25\penalty0
  (5):\penalty0 2298--2310, May 2016.
\newblock URL \url{https://doi.org/10.1109/TIP.2016.2541960}.

\bibitem[{Bian} et~al.(2014){Bian}, {Luo}, and {Huang}]{Bian2014}
{Bian}, S., {Luo}, W., and {Huang}, J.
\newblock Exposing fake bit rate videos and estimating original bit rates.
\newblock \emph{IEEE Transactions on Circuits and Systems for Video
  Technology}, 24\penalty0 (12):\penalty0 2144--2154, December 2014.
\newblock URL \url{https://doi.org/10.1109/TCSVT.2014.2334031}.

\bibitem[{Bianchi} \& {Piva}(2012){Bianchi} and {Piva}]{Bianchi2012}
{Bianchi}, T. and {Piva}, A.
\newblock Image forgery localization via block-grained analysis of jpeg
  artifacts.
\newblock \emph{IEEE Transactions on Information Forensics and Security},
  7\penalty0 (3):\penalty0 1003--1017, June 2012.
\newblock URL \url{https://doi.org/10.1109/TIFS.2012.2187516}.

\bibitem[Bird(2015)]{Bird2015}
Bird, M.
\newblock The video in which {Greece}'s finance minister gives {Germany} the
  finger has several bizarre new twists.
\newblock March 2015.
\newblock URL
  \url{https://www.businessinsider.com/yanis-varoufakis-middle-finger-controversy-real-fake-bohmermann-jauch-2015-3}.
\newblock (Accessed on 04/17/2019).

\bibitem[Brundage et~al.(2018)Brundage, Avin, Clark, Toner, Eckersley,
  Garfinkel, Dafoe, Scharre, Zeitzoff, Filar, Anderson, Roff, Allen,
  Steinhardt, Flynn, h\'{E}igeartaigh, Beard, Belfield, Farquhar, Lyle,
  Crootof, Evans, Page, Bryson, Yampolskiy, and Amodei]{Brundage2018}
Brundage, M., Avin, S., Clark, J., Toner, H., Eckersley, P., Garfinkel, B.,
  Dafoe, A., Scharre, P., Zeitzoff, T., Filar, B., Anderson, H., Roff, H.,
  Allen, G.~C., Steinhardt, J., Flynn, C., h\'{E}igeartaigh, S.~{\'{O}}.,
  Beard, S., Belfield, H., Farquhar, S., Lyle, C., Crootof, R., Evans, O.,
  Page, M., Bryson, J., Yampolskiy, R., and Amodei, D.
\newblock The malicious use of artificial intelligence: Forecasting,
  prevention, and mitigation.
\newblock \emph{arXiv:1802.07228v1}, February 2018.
\newblock URL \url{http://arxiv.org/abs/1802.07228v1}.

\bibitem[{Bulan} et~al.(2009){Bulan}, {Mao}, and {Sharma}]{Bulan2009}
{Bulan}, O., {Mao}, J., and {Sharma}, G.
\newblock Geometric distortion signatures for printer identification.
\newblock \emph{Proceedings of the IEEE International Conference on Acoustics,
  Speech and Signal Processing}, pp.\  1401--1404, April 2009.
\newblock URL \url{https://doi.org/10.1109/ICASSP.2009.4959855}.

\bibitem[Chesney \& Citron(2018)Chesney and Citron]{Chesney2018}
Chesney, R. and Citron, D.~K.
\newblock Disinformation on steroids: The threat of deep fakes.
\newblock October 2018.
\newblock URL
  \url{https://www.cfr.org/report/deep-fake-disinformation-steroids}.
\newblock (Accessed on 04/17/2019).

\bibitem[Cole(2018)]{Cole2018}
Cole, S.
\newblock Fake porn makers are worried about accidentally making child porn.
\newblock February 2018.
\newblock URL
  \url{https://motherboard.vice.com/en_us/article/evmkxa/ai-fake-porn-deepfakes-child-pornography-emma-watson-elle-fanning}.
\newblock (Accessed on 04/17/2019).

\bibitem[Curtis(2018)]{Curtis2018}
Curtis, C.
\newblock Deepfakes are being weaponized to silence women — but this woman is
  fighting back.
\newblock October 2018.
\newblock URL
  \url{https://thenextweb.com/code-word/2018/10/05/deepfakes-are-being-weaponized-to-silence-women-but-this-woman-is-fighting-back/}.
\newblock (Accessed on 04/17/2019).

\bibitem[{D'Amiano} et~al.(2015){D'Amiano}, {Cozzolino}, {Poggi}, and
  {Verdoliva}]{Amiano2015}
{D'Amiano}, L., {Cozzolino}, D., {Poggi}, G., and {Verdoliva}, L.
\newblock Video forgery detection and localization based on 3d {PatchMatch}.
\newblock \emph{Proceedings of the IEEE International Conference on Multimedia
  Expo Workshops}, pp.\  1--6, June 2015.
\newblock URL \url{https://doi.org/10.1109/ICMEW.2015.7169805}.
\newblock {Turin, Italy}.

\bibitem[{D'Amiano} et~al.(2019){D'Amiano}, {Cozzolino}, {Poggi}, and
  {Verdoliva}]{Amiano2019}
{D'Amiano}, L., {Cozzolino}, D., {Poggi}, G., and {Verdoliva}, L.
\newblock A {PatchMatch}-based dense-field algorithm for video copy–move
  detection and localization.
\newblock \emph{IEEE Transactions on Circuits and Systems for Video
  Technology}, 29\penalty0 (3):\penalty0 669--682, March 2019.
\newblock URL \url{https://doi.org/10.1109/TCSVT.2018.2804768}.

\bibitem[D'Avino et~al.(2017)D'Avino, Cozzolino, Poggi, and
  Verdoliva]{Avino2017}
D'Avino, D., Cozzolino, D., Poggi, G., and Verdoliva, L.
\newblock Autoencoder with recurrent neural networks for video forgery
  detection.
\newblock \emph{Proceedings of the IS\&T Electronic Imaging}, 2017\penalty0
  (7):\penalty0 92--99, January 2017.
\newblock URL \url{https://doi.org/10.2352/ISSN.2470-1173.2017.7.MWSF-330}.
\newblock {Burlingame, CA}.

\bibitem[{Fan} et~al.(2011){Fan}, {Kot}, {Cao}, and {Sattar}]{Fan2011}
{Fan}, J., {Kot}, A.~C., {Cao}, H., and {Sattar}, F.
\newblock Modeling the exif-image correlation for image manipulation detection.
\newblock \emph{Proceedings of the IEEE International Conference on Image
  Processing}, pp.\  1945--1948, September 2011.
\newblock URL \url{https://doi.org/10.1109/ICIP.2011.6115853}.
\newblock {Brussels, Belgium}.

\bibitem[{Guan} et~al.(2019){Guan}, {Kozak}, {Robertson}, {Lee}, {Yates},
  {Delgado}, {Zhou}, {Kheyrkhah}, {Smith}, and {Fiscus}]{Guan2019}
{Guan}, H., {Kozak}, M., {Robertson}, E., {Lee}, Y., {Yates}, A.~N., {Delgado},
  A., {Zhou}, D., {Kheyrkhah}, T., {Smith}, J., and {Fiscus}, J.
\newblock Mfc datasets: Large-scale benchmark datasets for media forensic
  challenge evaluation.
\newblock \emph{Proceedings of the IEEE Winter Applications of Computer Vision
  Workshops}, pp.\  63--72, January 2019.
\newblock URL \url{https://doi.org/10.1109/WACVW.2019.00018}.
\newblock {Waikoloa Village, HI}.

\bibitem[{G\"{u}era} \& {Delp}(2018){G\"{u}era} and {Delp}]{Guera2018b}
{G\"{u}era}, D. and {Delp}, E.~J.
\newblock Deepfake video detection using recurrent neural networks.
\newblock \emph{Proceedings of the IEEE International Conference on Advanced
  Video and Signal Based Surveillance}, pp.\  1--6, November 2018.
\newblock URL \url{https://doi.org/10.1109/AVSS.2018.8639163}.
\newblock {Auckland, New Zealand}.

\bibitem[Hsu et~al.(2008)Hsu, Hung, Lin, and Hsu]{Hsu2008}
Hsu, C.-C., Hung, T.-Y., Lin, C.-W., and Hsu, C.-T.
\newblock Video forgery detection using correlation of noise residue.
\newblock \emph{Proceedings of IEEE Workshop on Multimedia Signal Processing},
  pp.\  170--174, October 2008.
\newblock URL \url{https://doi.org/10.1109/MMSP.2008.4665069}.
\newblock {Cairns, Qld, Australia}.

\bibitem[Huh et~al.(2018)Huh, Liu, Owens, and Efros]{Huh2018}
Huh, M., Liu, A., Owens, A., and Efros, A.~A.
\newblock Fighting fake news: Image splice detection via learned
  self-consistency.
\newblock \emph{Proceedings of the European Conference on Computer Vision},
  pp.\  106--124, September 2018.
\newblock URL \url{https://doi.org/10.1007/978-3-030-01252-6_7}.
\newblock {Munich, Germany}.

\bibitem[{Iuliani} et~al.(2019){Iuliani}, {Shullani}, {Fontani}, {Meucci}, and
  {Piva}]{Iuliani2019}
{Iuliani}, M., {Shullani}, D., {Fontani}, M., {Meucci}, S., and {Piva}, A.
\newblock A video forensic framework for the unsupervised analysis of
  {MP4}-like file container.
\newblock \emph{IEEE Transactions on Information Forensics and Security},
  14\penalty0 (3):\penalty0 635--645, March 2019.
\newblock URL \url{https://doi.org/10.1109/TIFS.2018.2859760}.

\bibitem[Jack(2007)]{jack2007_ch13}
Jack, K.
\newblock Chapter 13 - {MPEG}-2.
\newblock In Jack, K. (ed.), \emph{Video Demystified: A Handbook for the
  Digital Engineer}, pp.\  577--737. Newnes, {Burlington, MA}, 2007.
\newblock URL \url{https://doi.org/10.1016/B978-075068395-1/50013-4}.

\bibitem[{Khanna} et~al.(2008){Khanna}, {Chiu}, {Allebach}, and
  {Delp}]{Khanna2008}
{Khanna}, N., {Chiu}, G. T.~., {Allebach}, J.~P., and {Delp}, E.~J.
\newblock Forensic techniques for classifying scanner, computer generated and
  digital camera images.
\newblock pp.\  1653--1656, March 2008.
\newblock URL \url{https://doi.org/10.1109/ICASSP.2008.4517944}.
\newblock {Las Vegas, NV}.

\bibitem[Korshunov \& Marcel(2018)Korshunov and Marcel]{Korshunov2018}
Korshunov, P. and Marcel, S.
\newblock Deepfakes: a new threat to face recognition? assessment and
  detection.
\newblock \emph{arXiv:1812.08685v1}, March 2018.
\newblock URL \url{https://arxiv.org/abs/1812.08685v1}.

\bibitem[{Korshunova} et~al.(2017){Korshunova}, {Shi}, {Dambre}, and
  {Theis}]{Korshunova2017}
{Korshunova}, I., {Shi}, W., {Dambre}, J., and {Theis}, L.
\newblock Fast face-swap using convolutional neural networks.
\newblock \emph{Proceedings of the IEEE International Conference on Computer
  Vision}, pp.\  3697--3705, October 2017.
\newblock URL \url{https://doi.org/10.1109/ICCV.2017.397}.
\newblock {Venice, Italy}.

\bibitem[{Lameri} et~al.(2017){Lameri}, {Bondi}, {Bestagin}, and
  {Tubaro}]{Lameri2017}
{Lameri}, S., {Bondi}, L., {Bestagin}, P., and {Tubaro}, S.
\newblock Near-duplicate video detection exploiting noise residual traces.
\newblock \emph{Proceedings of the IEEE International Conference on Image
  Processing}, pp.\  1497--1501, September 2017.
\newblock URL \url{https://doi.org/10.1109/ICIP.2017.8296531}.
\newblock {Beijing, China}.

\bibitem[{Li} et~al.(2018){Li}, {Chang}, and {Lyu}]{Li2018}
{Li}, Y., {Chang}, M., and {Lyu}, S.
\newblock In ictu oculi: Exposing {AI} created fake videos by detecting eye
  blinking.
\newblock \emph{Proceedings of the IEEE International Workshop on Information
  Forensics and Security}, pp.\  1--7, December 2018.
\newblock URL \url{https://doi.org/10.1109/WIFS.2018.8630787}.
\newblock {Hong Kong, China}.

\bibitem[{Mandelli} et~al.(2018){Mandelli}, {Bestagini}, {Tubaro}, {Cozzolino},
  and {Verdoliva}]{Mandelli2018}
{Mandelli}, S., {Bestagini}, P., {Tubaro}, S., {Cozzolino}, D., and
  {Verdoliva}, L.
\newblock Blind detection and localization of video temporal splicing
  exploiting sensor-based footprints.
\newblock \emph{Proceedings of the European Signal Processing Conference}, pp.\
   1362--1366, September 2018.
\newblock URL \url{https://doi.org/10.23919/EUSIPCO.2018.8553511}.
\newblock {Rome, Italy}.

\bibitem[{Matern} et~al.(2019){Matern}, {Riess}, and {Stamminger}]{Matern2018}
{Matern}, F., {Riess}, C., and {Stamminger}, M.
\newblock Exploiting visual artifacts to expose deepfakes and face
  manipulations.
\newblock \emph{Proceedings of the IEEE Winter Applications of Computer Vision
  Workshops}, pp.\  83--92, January 2019.
\newblock URL \url{https://doi.org/10.1109/WACVW.2019.00020}.
\newblock {Waikoloa Village, HI}.

\bibitem[McKinney(2010)]{Mckinney2010}
McKinney, W.
\newblock Data structures for statistical computing in python.
\newblock \emph{Proceedings of the Python in Science Conference}, pp.\  51--56,
  June 2010.
\newblock URL
  \url{http://conference.scipy.org/proceedings/scipy2010/mckinney.html}.
\newblock {Austin, TX}.

\bibitem[Milani et~al.(2012)Milani, Fontani, Bestagini, Barni, Piva,
  Tagliasacchi, and Tubaro]{Milani2012}
Milani, S., Fontani, M., Bestagini, P., Barni, M., Piva, A., Tagliasacchi, M.,
  and Tubaro, S.
\newblock An overview on video forensics.
\newblock \emph{APSIPA Transactions on Signal and Information Processing},
  1:\penalty0 e2, August 2012.
\newblock URL \url{https://doi.org/10.1017/ATSIP.2012.2}.

\bibitem[Mullan et~al.(2017)Mullan, Cozzolino, Verdoliva, and
  Riess]{Mullan2017}
Mullan, P., Cozzolino, D., Verdoliva, L., and Riess, C.
\newblock Residual-based forensic comparison of video sequences.
\newblock \emph{Proceedings of the IEEE International Conference on Image
  Processing}, pp.\  1507--1511, September 2017.
\newblock URL \url{https://doi.org/10.1109/ICIP.2017.8296533}.
\newblock {Beijing, China}.

\bibitem[Pedregosa et~al.(2011)Pedregosa, Varoquaux, Gramfort, Michel, Thirion,
  Grisel, Blondel, Prettenhofer, Weiss, Dubourg, Vanderplas, Passos,
  Cournapeau, Brucher, Perrot, and Duchesnay]{Pedregosa2011}
Pedregosa, F., Varoquaux, G., Gramfort, A., Michel, V., Thirion, B., Grisel,
  O., Blondel, M., Prettenhofer, P., Weiss, R., Dubourg, V., Vanderplas, J.,
  Passos, A., Cournapeau, D., Brucher, M., Perrot, M., and Duchesnay, E.
\newblock Scikit-learn: Machine learning in {P}ython.
\newblock \emph{Journal of Machine Learning Research}, 12:\penalty0 2825--2830,
  November 2011.
\newblock URL \url{http://dl.acm.org/citation.cfm?id=1953048.2078195}.

\bibitem[R\"{o}ssler et~al.(2018)R\"{o}ssler, Cozzolino, Verdoliva, Riess,
  Thies, and Nie{\ss}ner]{roessler2018}
R\"{o}ssler, A., Cozzolino, D., Verdoliva, L., Riess, C., Thies, J., and
  Nie{\ss}ner, M.
\newblock Faceforensics: A large-scale video dataset for forgery detection in
  human faces.
\newblock \emph{arXiv:1803.09179}, March 2018.
\newblock URL \url{https://arxiv.org/abs/1803.09179}.

\bibitem[Saito \& Rehmsmeier(2015)Saito and Rehmsmeier]{Saito2015}
Saito, T. and Rehmsmeier, M.
\newblock The precision-recall plot is more informative than the roc plot when
  evaluating binary classifiers on imbalanced datasets.
\newblock \emph{PLoS ONE}, 10\penalty0 (3):\penalty0 e0118432, March 2015.
\newblock URL \url{https://doi.org/10.1371/journal.pone.0118432}.

\bibitem[{Stamm} et~al.(2012{\natexlab{a}}){Stamm}, {Lin}, and
  {Liu}]{Stamm2012}
{Stamm}, M.~C., {Lin}, W.~S., and {Liu}, K. J.~R.
\newblock Forensics vs. anti-forensics: A decision and game theoretic
  framework.
\newblock \emph{Proceedings of the IEEE International Conference on Acoustics,
  Speech and Signal Processing}, pp.\  1749--1752, March 2012{\natexlab{a}}.
\newblock URL \url{https://doi.org/10.1109/ICASSP.2012.6288237}.
\newblock {Kyoto, Japan}.

\bibitem[{Stamm} et~al.(2012{\natexlab{b}}){Stamm}, {Lin}, and
  {Liu}]{Stamm2012b}
{Stamm}, M.~C., {Lin}, W.~S., and {Liu}, K. J.~R.
\newblock Temporal forensics and anti-forensics for motion compensated video.
\newblock \emph{IEEE Transactions on Information Forensics and Security},
  7\penalty0 (4):\penalty0 1315--1329, August 2012{\natexlab{b}}.
\newblock URL \url{https://doi.org/10.1109/TIFS.2012.2205568}.

\bibitem[Thies et~al.(2016)Thies, Zollh\"{o}fer, Stamminger, Theobalt, and
  Nie{\ss}ner]{Thies2016}
Thies, J., Zollh\"{o}fer, M., Stamminger, M., Theobalt, C., and Nie{\ss}ner, M.
\newblock Face2{F}ace: Real-time face capture and reenactment of {RGB} videos.
\newblock \emph{Proceedings of the IEEE Conference on Computer Vision and
  Pattern Recognition}, pp.\  2387--2395, June 2016.
\newblock URL \url{https://doi.org/10.1109/CVPR.2016.262}.
\newblock {Las Vegas, NV}.

\bibitem[Vincent(2018)]{Vincent2018}
Vincent, J.
\newblock {US} lawmakers say {AI} deepfakes ‘have the potential to disrupt
  every facet of our society’.
\newblock September 2018.
\newblock URL
  \url{https://www.theverge.com/2018/9/14/17859188/ai-deepfakes-national-security-threat-lawmakers-letter-intelligence-community}.
\newblock (Accessed on 04/17/2019).

\bibitem[{Zannettou} et~al.(2018){Zannettou}, {Chatzis}, {Papadamou}, and
  {Sirivianos}]{Zannettou2018}
{Zannettou}, S., {Chatzis}, S., {Papadamou}, K., and {Sirivianos}, M.
\newblock The good, the bad and the bait: Detecting and characterizing
  clickbait on youtube.
\newblock \emph{Proceedings of the IEEE Security and Privacy Workshops}, pp.\
  63--69, May 2018.
\newblock URL \url{https://doi.org/10.1109/SPW.2018.00018}.
\newblock {San Francisco, CA}.

\end{thebibliography}
\bibliographystyle{icml2019}

\end{document}